\documentclass{article} 
\usepackage{iclr2021_conference,times}

\usepackage{amsmath,amsfonts,bm,xspace}

\def\eqref#1{equation~\ref{#1}}

\def\1{\bm{1}}

\def\vp{{\bm{p}}}

\def\vx{{\bm{x}}}

\DeclareMathAlphabet{\mathsfit}{\encodingdefault}{\sfdefault}{m}{sl}
\SetMathAlphabet{\mathsfit}{bold}{\encodingdefault}{\sfdefault}{bx}{n}

\newcommand{\etc}{\textit{etc.}\xspace}
\newcommand{\ie}{\textit{i.e.,}\xspace}

\usepackage[colorlinks,linkcolor=blue]{hyperref}
\usepackage{url}
\usepackage{hhline}
\usepackage{subcaption}
\usepackage{xcolor}
\usepackage{wrapfig}
\usepackage{microtype}
\usepackage{graphicx}
\usepackage{booktabs} 
\usepackage{colortbl}
\usepackage{ dsfont }

\title{CSI: Enhancing the Robustness of 3D Point Cloud Recognition against Corruption}

\author{Zhuoyuan Wu$^{1,3}$ \thanks{equal contribution; work done when Zhuoyuan is an intern at University of Wisconsin, Madison} \quad  Jiachen Sun$^{2}$ \footnotemark[1] \quad  Chaowei Xiao$^{3}$ \\
    $^1$Peking University \quad   $^2$University of Michigan \quad $^3$University of Wisconsin, Madison\\
    \texttt{wuzhuoyuan@pku.edu.cn, jiachens@umich.edu, cxiao34@wisc.edu}
}

\iclrfinalcopy 
\begin{document}

\maketitle

\begin{abstract}
Despite recent advancements in deep neural networks for point cloud recognition, real-world safety-critical applications present challenges due to unavoidable data corruption. 
Current models often fall short in generalizing to unforeseen distribution shifts.
In this study, we harness the inherent set property of point cloud data to introduce a novel critical subset identification (CSI) method, aiming to bolster recognition robustness in the face of data corruption. Our CSI framework integrates two pivotal components: density-aware sampling (DAS) and self-entropy minimization (SEM), which cater to static and dynamic CSI, respectively. DAS ensures efficient robust anchor point sampling by factoring in local density, while SEM is employed during training to accentuate the most salient point-to-point attention. Evaluations reveal that our CSI approach yields error rates of 18.4\% and 16.3\% on ModelNet40-C and PointCloud-C, respectively, marking a notable improvement over state-of-the-art methods by margins of 5.2\% and 4.2\% on the respective benchmarks. Code is available at \href{https://github.com/masterwu2115/CSI/tree/main}{https://github.com/masterwu2115/CSI/tree/main}
\end{abstract}

\section{Introduction}
\label{sec:intro}

Point cloud recognition, the process of identifying and categorizing objects or scenes represented as unordered point sets in 3D space, has pivotal applications in fields such as autonomous driving~\citep{DBLP:conf/cvpr/ChenMWLX17, DBLP:conf/mir/YueWSKS18}, medical imaging, virtual and augmented reality~\citep{Garrido_Rodrigues_Augusto_Sousa_Jacob_Castro_Silva_2021}, and 3D printing~\citep{Tashi_Ullah_Kubo_2019}. However, real-world deployment presents challenges including noise and distortion arising from sensor errors, occlusions, and missing data, which can significantly impair the accuracy and robustness of recognition models~\citep{sun2022benchmarking, ren2022benchmarking}. For instance, LiDAR sensors, although essential to autonomous vehicles, are susceptible to adverse weather conditions, dirt accumulation, and hardware malfunctions. Similarly, in medical imaging, where point cloud data aids in 3D reconstructions from MRI or CT scans, the presence of artifacts, noise, and incomplete data — arising from limited resolution, patient movement, or implants — poses substantial challenges. Therefore, exploring avenues to enhance robustness against corrupted point cloud data is imperative.


To boost corruption robustness, the predominant strategy is data augmentation. Analogous to augmentation strategies applied to 2D images~\citep{Yun_Han_Chun_Oh_Yoo_Choe_2019, DBLP:conf/iclr/ZhangCDL18}, several methods are proposed to enhance training diversity and foster the learning of more robust features, including PointMixup~\citep{DBLP:conf/eccv/ChenHGMMYS20}, PointCutMix~\citep{Zhang_Chen_Ouyang_Liu_Zhu_Chen_Meng_Wu_2022}, RSMix~\citep{DBLP:conf/cvpr/LeeLLLLWL21}, and PointWOLF~\citep{DBLP:conf/iccv/KimLHLHK21}. However, as highlighted by~\citet{sun2022benchmarking}, different data augmentation techniques have varying degrees of effectiveness against distinct types of corruption. This is partially because data augmentation can inadvertently introduce noise and artifacts, thereby degrading the quality of the training dataset and potentially leading to overfitting and diminished generalization performance on unseen distribution shifts. Moreover, the process of augmenting data often relies on heuristic approaches, which may not always align with the underlying data distribution, thus not providing a principled manner to enhance robustness. 
In addition to data augmentation, \citet{sun2022benchmarking} illustrated that the robustness can also be improved from the model architecture perspective. By studying the robustness among various 3D architectures including PointNet~\citep{Charles_Su_Kaichun_Guibas_2017}, PointNet++~\citep{qi2017pointnet++}, DGCNN~\citep{wang2019dynamic}, \etc, they revealed that Transformers, specifically PCT~\citep{guo2021pct}, can significantly enhance the robustness of point cloud recognition. 

In this study, we address the robustness challenge inherent in point cloud recognition by harnessing its fundamental set property, a perspective that complements existing strategies. A typical point cloud in ModelNet~\citep{wu20153d} comprises between 1000 to 2000 points. However, humans require far fewer points to recognize a point cloud accurately. While the additional points can contribute to a richer semantic understanding, they also present opportunities for data corruption. Motivated by this observation, we introduce a novel critical subset identification (CSI) method, encompassing two distinct components that facilitate both static and dynamic CSI, thereby aiding in the robust recognition of point clouds.

\begin{figure}
  \centering
  \includegraphics[width=1\textwidth]{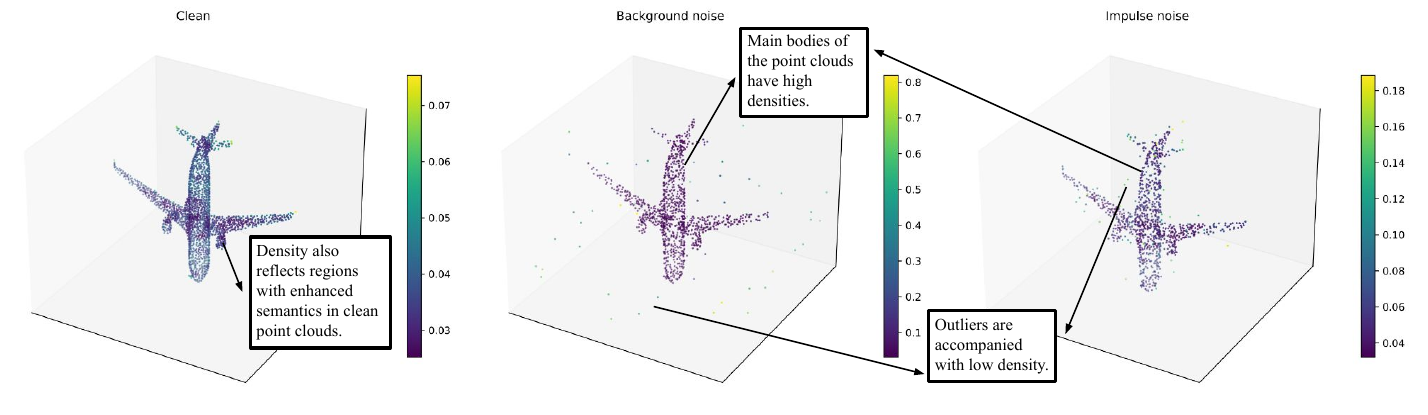}
  \caption{Illustration of Point Cloud Density. For each point, we calculate the mean distance of its $k$NN as the density score. Therefore, a smaller value denotes a larger density. }
  \label{fig:density}
  \vspace{-0.8cm}
\end{figure}

The first component of CSI is a novel anchor point sampling strategy to achieve static critical subset identification. Anchor point sampling is pivotal in point cloud models, facilitating local feature learning and enhancing efficiency. The strategy of Farthest point sampling (FPS) is widely adopted~\citep{Muzahid_Wan_Sohel_Wu_Hou_2021, guo2021pct, DBLP:conf/iclr/MaQYR022}. While FPS is adept at preserving the skeletal structure of the given point cloud, it overlooks the potential data corruption, rendering it vulnerable to noise and distortion. We note that outlier data, typically distanced from other points, tend to exhibit lower density. Drawing from this observation, we propose a density-aware sampling (DAS) technique to down-sample the point cloud in accordance with its density. By allocating higher probabilities to points with greater density, our approach ensures a more representative subset of points is retained, thereby better preserving the semantic information inherent in the point cloud. The inherent versatility of our approach introduces an additional layer of unpredictability, fortifying the robustness of the model against data corruption. 
The second component of CSI is a new optimization objective to achieve dynamic critical subset identification during training. Recent studies~\citep{DBLP:conf/iclr/WangSLOD21, DBLP:conf/nips/ZhangLF22} illustrate that employing batch-level entropy minimization in the logit space can bolster robustness during test time. Our proposition stems from the insight that entropy minimization in the embedding space could be instrumental for critical subset identification, as the objective is to enhance the saliency corresponding to the most critical point or region-level feature. By integrating the entropy minimization loss as a joint optimization objective alongside the classification objective, our method unfolds an unsupervised and architecture-agnostic solution. Through adept balancing of the entropy loss term and the classification objective, our approach cultivates enhanced robustness in confronting data corruption. 

We conduct an extensive evaluation of our proposed CSI method on two widely recognized corruption robustness benchmarks: ModelNet40-C and PointCloud-C. The experimental outcomes reveal that our CSI method achieves error rates of 18.4\% and 16.3\% on ModelNet40-C and PointCloud-C, markedly surpassing the state-of-the-art methods by \textbf{5.2}\% and \textbf{4.2}\% on the respective benchmarks. Additionally, we carry out thorough ablation studies to elucidate the effectiveness of CSI from various perspectives and dimensions.


\begin{figure}
  \centering
  \includegraphics[width=1\textwidth]{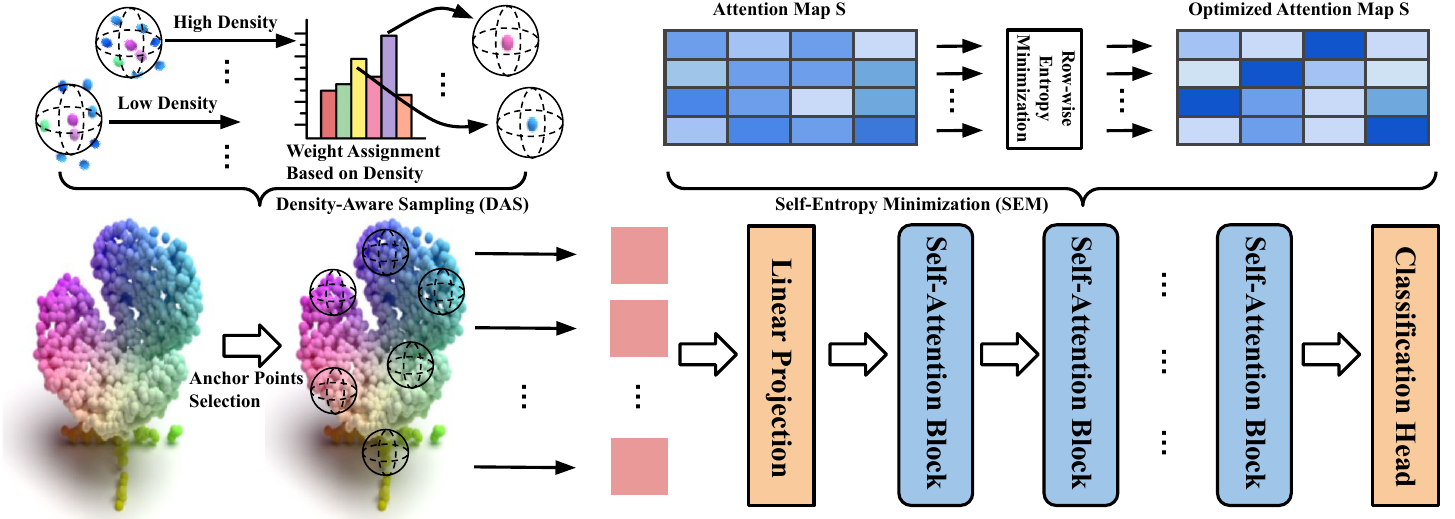}
  \caption{Overview of Our Critical Subset Identification (CSI) Method for Robust Point Cloud Recognition. CSI consists of two key components: density-aware sampling (DAS) and self-entropy minimization (SEM) objective, which achieve static and dynamic CSI, respectively. DAS ensures a robust anchor point sampling process and SEM increases the saliency of the critical correspondence between points/regions.}
  \label{fig:csi}
\end{figure}

\section{Background and Related Work}
In this section, we review a few topics that are related to our study, including deep learning models and robustness analysis for point clouds.
\label{sec:background}
\subsection{Deep Learning for Point Cloud Recognition}
Point cloud classification is a popular research topic in scene understanding. Recent years have witnessed vigorous development of this topic owing to deep learning. The seminal work of PointNet~\citep{Charles_Su_Kaichun_Guibas_2017} lays the foundation by employing a multi-layer perceptron (MLP) to distill local features from each point, coupled with a max-pooling layer to amalgamate these into a global feature vector for the classification task. This further serves as a general specification for follow-up point cloud recognition studies. Building upon this, PointNet++~\citep{qi2017pointnet++} extends the paradigm by introducing hierarchical feature learning through a series of nested local regions at varying scales. Inspired by these pioneering efforts, a slew of voxel-based methodologies such as RSCNN~\citep{Liu_Fan_Xiang_Pan_2019}, SimpleView~\citep{goyal2021revisiting}, and RPC~\citep{ren2022benchmarking} emerge, which discretize point clouds into 3D grids of voxels, subsequently employing convolution operations on these grids to extract features. Concurrently, graph-based approaches like DGCNN~\citep{wang2019dynamic}, ECC~\citep{DBLP:conf/cvpr/SimonovskyK17}, SPG~\citep{Landrieu_Simonovsky_2018}, and KCNet~\citep{DBLP:conf/cvpr/ShenFYT18} treat point clouds as graphs, leveraging graph convolutional networks to learn features. The narrative of innovation continues with the advent of attention-based methods such as PAConv~\citep{Xu_Ding_Zhao_Qi_2021}, and Transformer-based architectures like PCT~\citep{guo2021pct}, Point Transformer~\citep{Zhao_Jiang_Jia_Torr_Koltun_2021}, and Point Transformer V2~\citep{DBLP:conf/nips/0002LJLZ22}. Transformer architectures naturally align with point cloud data processing due to their order-invariance, obviating the need for positional encoding. These methodologies harness attention mechanisms to hone in on critical regions within the point cloud, achieving state-of-the-art performance in point cloud learning, and further enriching the landscape of point cloud classification.

\subsection{Robustness Analysis for Point Cloud Recognition} 
Numerous efforts have been directed towards enhancing the robustness of point cloud classifiers. For instance, Triangle-Net~\citep{Xiao_Wachs_2021} devises a feature extraction mechanism that remains invariant to positional, rotational, and scaling disturbances. While Triangle-Net exhibits remarkable robustness under severe corruption, its performance on clean data does not match the state-of-the-art benchmarks. On the other hand, PointASNL~\citep{Yan_Zheng_Li_Wang_Cui_2020} introduces adaptive sampling and local-nonlocal modules to bolster robustness, albeit with a limitation of processing a fixed number of points in its implementation.
Diverse strategies are explored in other works to bolster the adversarial robustness of a model. These include denoising and upsampling techniques~\citep{Wu_Zhou_Zhao_Yue_Keutzer_2019}, voting mechanisms on subsampled point clouds~\citep{DBLP:journals/corr/abs-2103-03046}, exploiting the relative positions of local features~\citep{Dong_Chen_Zhou_Hua_Zhang_Yu_2020}, adversarial training~\citep{sun2020adversarial} employing self-supervision~\citep{Sun_Cao_Choy_Yu_Anandkumar_Mao_Xiao_2021}, and diffusion models~\citep{sun2023critical}.
RobustPointSet~\citep{DBLP:journals/corr/abs-2011-11572} undertakes an evaluation of point cloud classifiers' robustness under various corruption scenarios. The findings reveal that basic data augmentations exhibit poor generalization capabilities when confronted with ``unseen'' corruptions. \citet{sun2022benchmarking} and \citet{ren2022benchmarking} introduced the ModelNet40-C and PointCloud-C benchmarks respectively, to facilitate a more comprehensive robustness analysis, each encompassing several types of corruption with multiple levels of severity. In this study, we utilize these benchmarks to conduct our evaluations.

\section{CSI: A Novel Critical Subset Identification Method}
\label{sec:csi}

In this section, we introduce the design of CSI, a new critical subset identification method for point cloud recognition. As noted in \S~\ref{sec:intro}, point clouds fundamentally constitute a set data structure, wherein a critical subset can be sufficiently representative for recognition. Figure~\ref{fig:csi} shows the overall workflow of CSI. As shown, CSI consists of two key components: density-aware sampling (DAS) and entropy minimization objective (SEM) that achieve static and dynamic critical subset identification during model training, respectively. CSI is also applicable to most point cloud recognition architectures and we detail their formulations below. 

\subsection{Density-Aware Sampling}
\label{sec:das}

In order to improve the competence of the models to recognize fine-grained patterns and generalize to complex scenes, methods like PointNet++~\citep{qi2017pointnet++}, DGCNN~\citep{wang2019dynamic}, PCT~\citep{guo2021pct} integrate neighbor embedding, trying to capture local structures via aggregating point features in the vicinity of each point. As mentioned in \S~\ref{sec:background}, such neighbor embedding usually comprises both sampling and grouping steps to improve the computation efficiency. Existing studies leverage random (RS) or farthest point sampling (FPS) to select anchor points for the grouping stage. Specifically, RS, the most intuitive method, downsamples the point cloud following a uniform distribution, while FPS aims to extract the most distant point with regard to the selected point iteratively. The grouping step aggregates the features into a group, leveraging $k$ nearest neighbor ($k$NN) or ball query, to achieve local feature learning. 

However, we find that both sampling methods exhibit sub-optimal performance. During the sampling phases of both training and inference stages, the inherent randomness in RS fails to ensure that the anchor points accurately represent the overarching skeleton of the object shape, resulting in diminished performance on clean inputs (detailed in \S~\ref{sec:evaluation}). Conversely, while FPS offers improved performance on clean data, it is vulnerable to exploitation through outlier points, a phenomenon frequently observed in corrupted point clouds~\citep{sun2022benchmarking,ren2022benchmarking}.
Motivated by these observations, we propose a simple yet effective strategy for robust anchor sampling: density-aware sampling (DAS). Our approach initially allocates a computed density weight to each point within the point cloud, followed by executing a weighted random sampling. Specifically, for each point $\vp_i \in \mathbb{R}^3$, we first identify its $k$NN point set $\mathcal{S}_i$ and calculate the averaged $\ell_2$ distance between $\vp_i$ and its $\mathcal{S}_i$ and further set the average distance as the threshold $t$:
\begin{equation}
    d_i = \frac{1}{k} \sum_{j \in \mathcal{S}_i} ||\vp_i - \vp_{j} ||_2, \quad t = \frac{1}{N} \sum_{i=1}^{N} d_i.
\end{equation}
The sampling weight is determined by: 
\begin{equation}
w_i = \sum_{j \in \mathcal{S}_i} \1[||\vp_i - \vp_{j} ||_2 < t], \quad w_i = \frac{w_i}{\sum_{j=1}^{N} w_j}.
\end{equation}
The underlying design philosophy of DAS is that the local density of a point usually bears a positive correlation with its significance in the point cloud, thereby facilitating a more representative sampling. This is particularly pronounced in the context of corrupted point clouds~\citep{sun2022benchmarking,ren2022benchmarking}. The incorporation of weighted randomness allows for the selection of low-density points during both training and inference phases, enhancing the diversity of the sample pool with high efficiency. Moreover, the application of hard thresholding effectively eliminates extreme outliers from being considered anchors, thereby maintaining the integrity of the data.

\subsection{Self-Entropy Minimization}
\label{sec:SEM}

In this section, we explore a fresh avenue of applying the entropy minimization objective, leveraging it for dynamic critical subset identification. As underscored in Section~\ref{sec:background}, Transformer architectures~\citep{vaswani2017attention} are inherently suited for 3D point cloud comprehension. We select PCT~\citep{guo2021pct} as the focal architecture in our study, given its incorporation of standard self-attention modules which offer notable transferability. Furthermore, research by~\citet{sun2022benchmarking} demonstrates the potential of PCT in maintaining promising performance under data corruption.
Formally, given a point cloud\footnote{We use a 2D tensor with a shape of $N \times 3$ to represent the point cloud set, so permutations in the first dimension will not alter $\mathbf{P}$ in practice.} $\mathbf{P} \in \mathbb{R}^{N\times 3}$, the input embedding and sampling and grouping layers (\S~\ref{sec:das}) project the point cloud from the coordinate space to the feature space to obtain the embedding feature $\mathbf{F}_s \in \mathbb{R}^{M\times D}$. Then four cascaded attention layers are followed to augment the down-sampled point features. Concretely, the procedure can be expressed as below:
\begin{align}
    \mathbf{F}_1   & = \mathrm{SA}^{1}\left(\mathbf{F}_s\right), \quad \mathbf{F}_{i} = \mathrm{SA}^{i}\left(\mathbf{F}_{i-1}\right),\quad i=2,3,4, \nonumber                                  \\
    \mathbf{F}_o   & = \mathrm{concat}\left( \mathbf{F}_1, \mathbf{F}_2, \mathbf{F}_3, \mathbf{F}_4\right)\cdot \mathbf{W}_o,
\end{align}
where $\mathrm{SA}^i$ represents the $i$-th self-attention layer, each having the same output dimension as its input. Specifically, let $\mathbf{Q} \in \mathbb{R}^{M\times d_a}, \mathbf{K} \in \mathbb{R}^{M\times d_a}, \mathbf{V} \in \mathbb{R}^{M\times d_e}$ be the query, key, and value matrices, respectively, generated by the linear transformation of the input features. The self-attention map $\mathbf{S} \in \mathbb{R}^{M\times M}$ and output $\mathbf{A} \in  \mathbb{R}^{M\times d_e} $ are calculated by
\begin{align}
    \mathbf{S} = \frac{\mathbf{Q} \mathbf{K}^\top}{\sqrt{d_a}}, \quad \mathbf{A} = \text{softmax}\left(\mathbf{S}\right)\mathbf{V}
\end{align}
$\mathbf{W}_o$ is the weight of a linear layer. Before transmitting to the classification head, a max-pooling layer is applied to extract the effective global feature $\mathbf{F}_g = \text{MaxPooling}\left(\mathbf{F}_o\right)$. Finally, the global feature is fed into an MLP to obtain the classification logits $\vx$.
\textbf{SEM for Self-Attention Modules}.  As mentioned in \S~\ref{sec:intro}, our goal is to highlight the critical subset within any input point clouds, which can be learned in the training stage. Entropy minimization is previously employed in test-time adaptation~\citep{DBLP:conf/iclr/WangSLOD21} to enhance corruption robustness, aiming to bolster the confidence of the most prominent logit at the batch level, a strategy validated as optimal for fully test-time adaptation by~\citet{goyal2022test}.
In this study, we propose to leverage the entropy minimization objective (SEM) for dynamic CSI during model training for robustness enhancement. The attention map, denoted as $\mathbf{S}$, is inherently aligned with our objective, encapsulating the strength of point-to-point correlations. Applying SEM to the row-wise embeddings present in $\mathbf{S}$ essentially amplifies the saliency of the most pivotal point-level feature corresponding to feature row $i$. Furthermore, row-level joint optimization can circumvent trivial solutions, fostering a more nuanced approach to achieving our goal. Specifically, our entropy minimization objective minimizes the Shannon entropy \citep{shannon2001mathematical} $\mathcal{H}(\mathbf{x}) = - \sum_i p_i \log p_i$ of the selected attention maps. Following~\citet{goyal2022test}, we utilize a temperature hyperparameter $\tau$ in SEM:
\begin{equation}
    \mathcal{H}\left(\mathbf{S}_{i,:}\right) = -\sum_{j=1}^M \frac{\exp\left(s_{i,j}/\tau\right)}{\sum_p \exp\left(s_{i,p}/\tau\right)} \log \left(\frac{\exp\left(s_{i,j}/\tau\right)}{\sum_p \exp\left(s_{i,p}/\tau\right)}\right), \quad    \mathcal{L}_\text{SEM} = \frac{1}{L} \frac{1}{M} \sum_{l=1}^L \sum_{i=1}^M \mathcal{H}\left(\mathbf{S}_{i,:}^l\right),
\end{equation}
where $L$ is the amount of the specified layers applying entropy minimization loss. 


\textbf{SEM for Other Models}. As outlined in Section~\ref{sec:background}, the general specification for point cloud recognition involves point-level feature learning, followed by the application of a symmetric function—such as a pooling operation—to extract global features for the prediction head. Consequently, the feature map preceding this symmetric function encapsulates the most salient point-level features. We thus apply channel-wise SEM to this layer. Similarly, given a feature map $\mathbf{F} \in \mathbb{R}^{M\times d_g}$, where $M$ is the number of point-level features and $d_g$ is the number of channels, the loss term is defined as:
\begin{equation}
     \mathcal{H}\left(\mathbf{F}_{:,i}\right) = -\sum_{j=1}^M \frac{\exp\left(f_{j,i}/\tau\right)}{\sum_p \exp\left(f_{p,i}/\tau\right)} \log \left(\frac{\exp\left(f_{j,i}/\tau\right)}{\sum_p \exp\left(f_{p,i}/\tau\right)}\right), \quad \mathcal{L}_\text{SEM} = \frac{1}{d_g} \sum_{i=1}^{d_g} \mathcal{H}\left(\mathbf{F}_{:,i}\right)
\label{eq:sem_general}
\end{equation}
This objective is designed to jointly amplify the saliency of the most pivotal feature accompanied by the most critical point along each channel.

The optimization process is carried out end-to-end, integrating the cross-entropy loss $\mathcal{L}_\text{CE}$ to formulate the loss function as $\mathcal{L} = \mathcal{L}_\text{CE} + \lambda \mathcal{L}_\text{SEM}$, where $\lambda$ serves as a tunable hyperparameter to harmonize the two loss components. It is also worth noting that SEM is an unsupervised task so there is no need for extra annotations. 

\section{Experiments and Results }
\label{sec:evaluation}

\subsection{Experimental Setups}

\textbf{Benchmarks}. We perform our experiments on the two widely recognized benchmarks for point cloud classification, ModelNet40-C~\citep{sun2022benchmarking} and PointCloud-C~\citep{ren2022benchmarking}, to evaluate the robustness of the proposed mechanisms. These benchmarks are derivatives of the foundational ModelNet40 dataset~\citep{wu20153d}, each incorporating distinct types of corruption to challenge trained models under evaluation. Note that the two benchmarks are only used in test time and the training set follows the official split of the ModelNet40. Specifically, ModelNet40-C contains 15 types of common corruption, which can be categorized into density, noise, and transformation types, while PointCloud-C contains 7 types of corruption, including jittering, global/local dropping, global/local adding, scaling, and rotation. Each corruption in both datasets has 5 severity levels.

\textbf{Metrics}. Following \citet{sun2022benchmarking}, we use the error rate (ER) as the metric for both ModelNet40-C and PointCloud-C. Specifically, $\text{ER} = 1 - \text{OA}$, where OA is the overall accuracy of corresponding clean/corruption data. We denote $\text{ER}_{\text{clean}} ^ f$ as the error rate for the classifier $f$ on the clean dataset (i.e., ModelNet40), $\text{ER}_{s,c} ^ f$ as the error rate for $f$ on corruption $c$ with severity $s$. The average error rate of certain corruption is defined as $\text{ER}_{c} ^ f = \frac{1}{5} \sum_{s=1}^5 \text{ER}_{s,c}^f$ and the overall average error rate is $\text{ER}_{\text{cor}}^f = \frac{1}{C}\sum_{c=1}^C \text{ER}_c^f$ where $C$ is the amount of corruption types (15 for ModelNet40-C and 7 for PointCloud-C).

\textbf{Implementation Details}. As mentioned in \S~\ref{sec:csi}, we use the representative Transformer architecture, PCT, in our main evaluation and additionally choose nine models to demonstrate the generalization of our proposed CSI, including PointNet~\citep{Charles_Su_Kaichun_Guibas_2017}, PointNet++~\citep{qi2017pointnet++}, DGCNN~\citep{wang2019dynamic}, RSCNN~\citep{Liu_Fan_Xiang_Pan_2019}, SimpleView~\citep{goyal2021revisiting}, GDANet~\citep{Xu_Zhang_Zhou_Xu_Qi_Qiao_2022}, CurveNet~\citep{Muzahid_Wan_Sohel_Wu_Hou_2021}, PAConv~\citep{Xu_Ding_Zhao_Qi_2021}, and RPC~\citep{ren2022benchmarking}. We adopt the same training recipe for all models for a fair comparison. Smoothed cross entropy~\citep{wang2019dynamic} is selected as the classification loss. During training, each point cloud samples 1024 points and we use the Adam optimizer \citep{Kingma_Ba_2014}. Each model is trained for 300 epochs and the best checkpoint is selected for evaluation. We implement our DAS based on 
\texttt{torch.multinomial}~\citep{DBLP:conf/nips/PaszkeGMLBCKLGA19}, where the hyperparameter $k=5$. All the experiments are performed on a single Tesla V100 GPU.

\subsection{Result on Popular Corruption Benchmarks}

\begin{wraptable}{r}{8cm}
\setlength\tabcolsep{8pt}
  \caption{ Error Rates of Different Model Architectures on ModelNet40-C and PointCloud-C.}
  \label{tb:main_result}
  \centering
\resizebox{\linewidth}{!}{
  \begin{tabular}{lccc}
  \noalign{\global\arrayrulewidth1pt}\hline\noalign{\global\arrayrulewidth0.4pt}
    & ModelNet40 & ModelNet40-C & PointCloud-C\\
    \cline{2-4}
    Model (\%) $\downarrow$ & $\text{ER}_\text{clean}$ & $\text{ER}_\text{cor}$ & $\text{ER}_\text{cor}$\\
\noalign{\global\arrayrulewidth1pt}\hline\noalign{\global\arrayrulewidth0.4pt}     
PointNet    &9.3  & 28.3 & 34.2   \\
PointNet++  &7.0  & 23.6 & 24.9   \\
DGCNN       &7.4  & 25.9 & 23.6   \\
RSCNN       &7.7  & 26.2 & 26.1   \\
SimpleView  &\textbf{6.1}  & 27.2 & 24.3   \\
GDANet      &7.5  & 25.6 & 21.1   \\
CurveNet    &6.6  & 22.7 & 22.1   \\
PAConv      &6.4  & -  & 27.0   \\
RPC         &7.0  & 26.3  & 20.5       \\
PCT         &7.1  & 25.5 & 21.9   \\
\rowcolor{gray!20} PCT+SEM & 7.2 & 21.7 & 20.9 \\
\rowcolor{gray!20} PCT+CSI & 7.3  & \textbf{18.4} & \textbf{16.3}   \\
\noalign{\global\arrayrulewidth1pt}\hline\noalign{\global\arrayrulewidth0.4pt}
  \end{tabular}
  }
\end{wraptable}
In this section, we introduce the evaluation results of our proposed CSI on two existing corruption benchmarks: ModelNet40-C and PointCloud-C. As mentioned before, we mainly focus on PCT and we default to integrate SEM into all self-attention layers and replace FPS with DAS in the original neighbor embedding layers. On both benchmarks, we use the nine representative models with different designs described above as standalone architectures for comparisons.
Details are described in the following.

\begin{figure}
  \centering
  \includegraphics[width=1\textwidth]{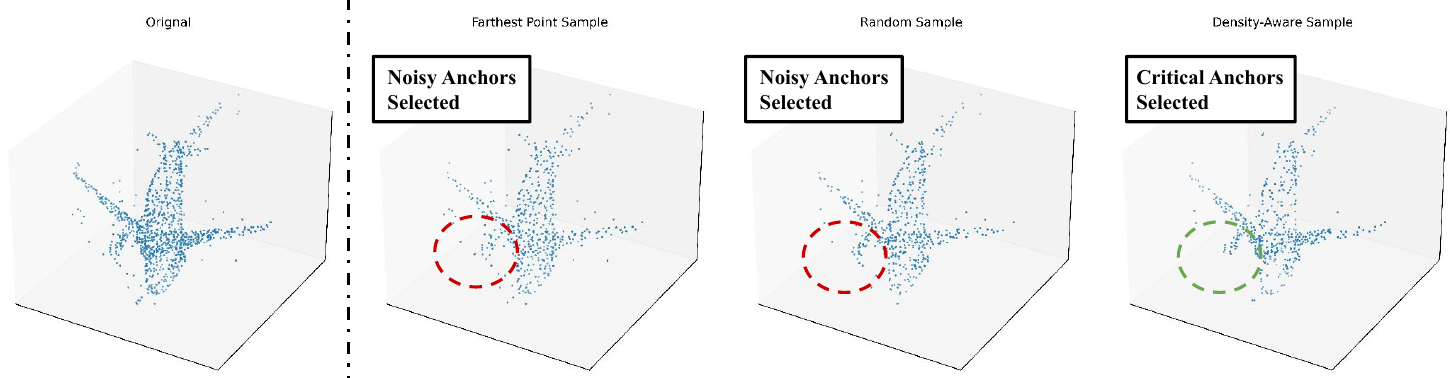}
  \caption{Example of Density-Aware Sampling (DAS). The left-most point cloud is the original one and the three sampled point clouds are selected from different sampling strategies. Our DAS effectively samples the most representative anchors statically.}
  \label{fig:das}
\end{figure}

\begin{wraptable}{l}{9cm}
\setlength\tabcolsep{8pt}
  \caption{Error Rates of Different Model Architectures on ModelNet40-C with Different Data Augmentation Strategies.
  }
  \label{tb:aug}
  \centering
\resizebox{\linewidth}{!}{
  \begin{tabular}{lccccc}
  \noalign{\global\arrayrulewidth1pt}\hline\noalign{\global\arrayrulewidth0.4pt}
    & Plain &\multicolumn{4}{c}{PointCutMix-R}\\
    \cline{2-6}
    Model (\%) $\downarrow$ & $\text{ER}_\text{cor}$ & $\text{ER}_\text{cor}$ & Density & Noise & Trans. \\
\noalign{\global\arrayrulewidth1pt}\hline\noalign{\global\arrayrulewidth0.4pt}     
PointNet   &28.3 & 21.8 & 30.5 & 18.0 & 16.9 \\
PointNet++  &23.6 & 19.1 & 28.1 & 12.2 & 17.0 \\
DGCNN   &25.9    & 17.3 & 28.9 & 11.4 & 11.5 \\
RSCNN   &26.2    & 17.9 & 25.0 & 13.0 & 15.8 \\
SimpleView  &27.2 & 19.7 & 31.2 & 11.3 & 16.5 \\
PCT     &25.5    & 16.3 & 27.1 & 10.5 & \textbf{11.2} \\
\rowcolor{gray!20} PCT+CSI   & \textbf{18.4} & \textbf{15.9} &\textbf{27.0} &\textbf{9.6} &\textbf{11.2} \\
\noalign{\global\arrayrulewidth1pt}\hline\noalign{\global\arrayrulewidth0.4pt}
  \end{tabular}
}
\end{wraptable}
\textbf{Performance on ModelNet40-C}. As shown in Table~\ref{tb:main_result}, our CSI-enhanced PCT achieves the best recognition performance on two corruption benchmarks, while maintaining high performance on the clean inputs. Specifically, DAS can greatly help with noise and density corruption types.  
We visualized the effect of DAS on certain corruptions in Figure~\ref{fig:das}, which clearly demonstrates its effectiveness. After incorporating SEM, we observe a marked performance boost, with the CSI-enhanced PCT outperforming the standard PCT on ModelNet40-C by approximately 7.1\%. 
Additionally, we ventured into experiments with data augmentation techniques. Following the insights from~\citet{sun2022benchmarking}, who demonstrated the efficacy of PointCutMix-R in tackling data corruption, we incorporated it into our study. The results were promising; our CSI mechanism further bolstered corruption robustness, registering an improvement of 0.4\%.  

\textbf{Performance on PointCloud-C}. Table~\ref{tb:main_result} also exhibits the robustness evaluation on PointCloud-C. Our CSI consistently achieves the best performance on PointCloud-C, attaining a significant improvement of 4.2\% compared to the previous state-of-the-art, especially in Jitter and Add-G. 
Besides, we also compare with several pre-trained methods OcCo~\citep{Wang_Liu_Yue_Lasenby_Kusner_2021} and Point-BERT~\citep{yu2021pointbert}, and data augmentation PointMixup~\citep{Zhang_Chen_Ouyang_Liu_Zhu_Chen_Meng_Wu_2022}, RSMix~\citep{Lee_Lee_Lee_Lee_Lee_Woo_Lee_2021}, PointWOLF~\citep{DBLP:conf/iccv/KimLHLHK21}, and WOLFMix~\citep{ren2022benchmarking}. Details can be found in Appendix~\ref{ap:eval}. The result shows that augmentation works well combined with our proposed CSI.

\begin{table}[!t]
\begin{minipage}{0.55\linewidth}
\centering
\setlength\tabcolsep{20pt}
\caption{ Ablation study on SEM Objective Imposed on Different Self-Attention (SA) Layers in PCT. $\dagger$ denotes the number of SA layers where SEM is applied.}
\label{tb:entropy loss layers}
\resizebox{\linewidth}{!}{
\begin{tabular}{lcc}
\noalign{\global\arrayrulewidth1pt}\hline\noalign{\global\arrayrulewidth0.4pt}
Model (\%) $\downarrow$ & $\text{ER}_\text{clean}$ & $\text{ER}_\text{cor}$ \\
\hline
PCT & 7.1 & 25.5 \\
+ SEM 1$\dagger$ & 7.7 & 22.0\\ 
+ SEM 2$\dagger$ & 7.9 & 22.1\\
+ SEM 3$\dagger$ & 7.5 & \textbf{21.5} \\
\rowcolor{gray!20}+ SEM 4$\dagger$ & \textbf{7.2} & 21.7 \\
\noalign{\global\arrayrulewidth1pt}\hline\noalign{\global\arrayrulewidth0.4pt}
\end{tabular}
}
\end{minipage}
\hfill
\begin{minipage}{0.43\linewidth}
\centering
\setlength\tabcolsep{16pt}
\caption{ Ablation Study on $\lambda$ in the Overall Loss Term. SEM is applied on all self-attention (\ie 4) layers in PCT.}
\label{tb:loss weight}
\resizebox{\linewidth}{!}{
\begin{tabular}{lcc}
\noalign{\global\arrayrulewidth1pt}\hline\noalign{\global\arrayrulewidth0.4pt}
$\lambda$ (\%) $\downarrow$ & $\text{ER}_\text{clean}$ & $\text{ER}_\text{cor}$ \\
\hline
1 & 9.1 & 23.5 \\
0.5 & 8.2 & 22.1\\
0.25 & 7.9 & 21.9\\
\rowcolor{gray!20} 0.1 & \textbf{7.2} & \textbf{21.7} \\
0.05 & 7.4 & 22.2 \\
\noalign{\global\arrayrulewidth1pt}\hline\noalign{\global\arrayrulewidth0.4pt}
\end{tabular}
}
\end{minipage}
\end{table}


\subsection{Comprehensive Ablation Study}
As introduced in \S~\ref{sec:csi}, both DAS and SEM are independent of the model architecture. In this section, we give a thorough analysis of our proposed modules. All experiments are down on ModelNet40-C.

\textbf{SEM on Different Self-Attention Layers}. In this experiment, we investigate which layers to apply our SEM objective can obtain the best gain. Specifically, we add the SEM loss in the last several layers. As shown in Table~\ref{tb:entropy loss layers}, imposing SEM objective in all self-attention layers accomplishes the best trade-off between $\text{ER}_{\text{clean}}$ and $\text{ER}_{\text{cor}}$.

\textbf{Weight $\lambda$ for SEM Loss}. In this experiment, we analyze the impact of our SEM weight $\lambda$ by designating a set of values \{0.05, 0.1, 0.25, 0.5, 1\}. We choose PCT as the baseline model and our SEM is applied in all self-attention layers, which is our default setting. As delineated in Table~\ref{tb:loss weight}, selecting an appropriate value for $\lambda$ is imperative to harmonize the two loss terms effectively, thereby realizing the most substantial enhancement in performance. While there exists a degree of fluctuation concerning the magnitude of improvement across different settings, the incorporation of our SEM objective continues to consistently bolster corruption robustness by a discernible margin. This consistency underscores the robustness of our SEM objective, demonstrating its capacity to contribute positively towards improving the resilience of the model against data corruption.

\textbf{SEM for Different Architectures}. In this ablation study, we extend the application of SEM across a variety of architectural designs including PointNet, PointNet++, DGCNN, RSCNN, SimpleView, and CurveNet. As elucidated in \S~\ref{sec:SEM}, we formulate a generalized SEM objective tailored to the specifications of point cloud recognition. To delve deeper into its effectiveness, a comparative analysis is carried out between our method, as expressed in Equation~\ref{eq:sem_general}, and SEM applied to the pooled global feature (i.e., $\textbf{F}_g$ in \S~\ref{sec:SEM}). The comparative results, presented in Table~\ref{tb:em on global feature}, underscore the consistent and significant enhancements our SEM objective brings about over the baseline models and other SEM objectives targeting the global feature, which is +1.34\% on average. The improvement is especially pronounced in SOTA architectures. This favorable outcome substantiates our hypothesis, affirming the validity of identifying critical subsets as a robust strategy for enhancing recognition performance, especially in scenarios prone to data corruption.

\begin{wraptable}{l}{8cm}
\centering
\setlength\tabcolsep{11pt}
\caption{ Ablation Study on SEM Imposed on Point-level (Equation~\ref{eq:sem_general}) and Global Feature Maps with Different Architectures.}
\label{tb:em on global feature}
\resizebox{\linewidth}{!}{

\begin{tabular}{lcc}
\noalign{\global\arrayrulewidth1pt}\hline\noalign{\global\arrayrulewidth0.4pt}
Model (\%) $\downarrow$ & $\text{ER}_\text{clean}$ & $\text{ER}_\text{cor}$ \\
\hline
PointNet & \textbf{9.3} & 28.3 \\
+SEM - Global Feature & 9.3 & 28.0\\
\rowcolor{gray!20}+SEM - Point-level Feature &9.4 &\textbf{27.1}\\
PointNet++ & 6.9 & 26.7 \\
+SEM - Global Feature& 6.9 & 26.4 \\
\rowcolor{gray!20}+SEM - Point-level Feature &\textbf{6.9} &\textbf{26.3}\\
DGCNN & 7.4 & 25.9\\
+SEM - Global Feature& 7.3 & 25.4 \\
\rowcolor{gray!20}+SEM - Point-level Feature &\textbf{7.3} &\textbf{23.9}\\
RSCNN & \textbf{7.7} & 29.2 \\
+SEM - Global Feature& 7.7 & 26.6 \\
\rowcolor{gray!20}+SEM - Point-level Feature &7.8 &\textbf{25.6}\\
CurveNet & 6.9 & 24.3 \\
+SEM - Global Feature& 6.9 & \textbf{22.9}\\
\rowcolor{gray!20}+SEM - Point-level Feature &\textbf{6.8} & 23.2\\
\noalign{\global\arrayrulewidth1pt}\hline\noalign{\global\arrayrulewidth0.4pt}
\end{tabular}
}
\end{wraptable}

\begin{table}[!t]
\begin{minipage}{0.65\linewidth}
\centering
\caption{ Ablation Study on Different Anchor Sampling Strategies.}
\label{tb:sampling strategy}
\begin{tabular}{lcc}
\noalign{\global\arrayrulewidth1pt}\hline\noalign{\global\arrayrulewidth0.4pt}
Anchor Sampling Strategy (\%) $\downarrow$ & $\text{ER}_\text{clean}$ & $\text{ER}_\text{cor}$ \\
\hline
Farthest Point Sampling & \textbf{7.2} & 21.7 \\
Random Sampling & 7.7 & 19.6 \\
\rowcolor{gray!20} Density-Aware Sampling & 7.3 & \textbf{18.4} \\
\noalign{\global\arrayrulewidth1pt}\hline\noalign{\global\arrayrulewidth0.4pt}
\end{tabular}
\end{minipage}
\hfill
\begin{minipage}{0.34\linewidth}
\centering
\caption{ Ablation study on $k$NN in Density-Aware Sampling (DAS).}
\label{tb:KNN density sampling}
\begin{tabular}{lcc}
\noalign{\global\arrayrulewidth1pt}\hline\noalign{\global\arrayrulewidth0.4pt}
$k$ (\%) $\downarrow$ &  $\text{ER}_\text{clean}$ & $\text{ER}_\text{cor}$ \\
\hline
\rowcolor{gray!20} 5 & \textbf{7.3} & \textbf{18.4} \\
10 & 7.5 & 18.6 \\
15 & 7.7 & 18.8\\
20 & 7.6 &  19.1\\
\noalign{\global\arrayrulewidth1pt}\hline\noalign{\global\arrayrulewidth0.4pt}
\end{tabular}
\end{minipage}
\end{table}

\textbf{Different Sampling Strategies}. We ablate the impacts of the sampling strategy in the neighbor embedding layers in this experiment. Our baseline model is PCT with SEM integrated. FPS serves as the sampling strategy in the original implementation of the PCT. Random Sampling (RS) can be viewed as a special case of DAS wherein all points are assumed to have equal density values. The evaluation of these strategies is encapsulated in Table~\ref{tb:sampling strategy}. A key observation is that when the density of points is taken into consideration, noisy points are less likely to be sampled, which elucidates why both RS and DAS exhibit superior performance in scenarios with noise corruption. Moreover, when compared to RS, DAS demonstrates enhanced performance concerning clean inputs, thereby showcasing its potential as a more robust sampling strategy that can adeptly handle both clean and corrupted data scenarios.

\textbf{$k$NN in DAS}. \S~\ref{sec:das} provides an implementation outline concerning the computation of density. The choice of value $k$ in $k$NN is pivotal in determining the threshold for DAS. Generally, a larger value of $k$ corresponds to a looser threshold, which may not effectively filter out outlier noisy points, as illustrated in Table~\ref{tb:KNN density sampling}. After a thorough evaluation, we opt for $k=5$ in our default DAS configuration.


\textbf{Different Definition for Density}. Apart from the density score delineated in \S~\ref{sec:das}, we propose two alternative definitions. Firstly, density can be characterized as the median distance of all mean distances computed by $k$NN. The distinction between this and the approach in \S~\ref{sec:das} lies in the employment of the $\ell_0$ or $\ell_1$ norm. Secondly, concerning the calculation of the threshold, alternatives such as the ball query method can be employed in lieu of $k$NN. In our experimentation, we adopted the ball query method with a radius of 0.1, aligning with the implementation in~\citet{qi2017pointnet++}. However, after a thorough evaluation, the implementation detailed in \S~\ref{sec:das} was selected for its superior performance in enhancing the robustness and maintaining the clean performance.

\textbf{DAS for Different Architectures}. The concept of neighbor embedding is not confined to PCT. In this analysis, we extend our exploration to two additional models, namely CurveNet and PointMLP~\citep{DBLP:conf/iclr/MaQYR022}, which also incorporate neighbor embedding layers. In these methods, we substitute the FPS strategy with our DAS across all neighbor embedding layers. The outcomes of this substitution are documented in Table~\ref{tb:DWS different architectures}. The integration of DAS markedly enhances the robustness of these models, thereby showcasing the potential of DAS as a robust sampling strategy across different architectural frameworks.

\begin{table}[!t]
\begin{minipage}{0.53\linewidth}
\centering
\caption{ Ablation Study on Density Definition.}
\label{tb:density definition}
\begin{tabular}{lcc}
\noalign{\global\arrayrulewidth1pt}\hline\noalign{\global\arrayrulewidth0.4pt}
Method (\%) $\downarrow$ & $\text{ER}_\text{clean}$ & $\text{ER}_\text{cor}$ \\
\hline
PCT+SEM & \textbf{7.2} & 21.7 \\
\rowcolor{gray!20}+ DAS - $k$NN $\ell_0$ & 7.3 & \textbf{18.4} \\
+ DAS - $k$NN $\ell_1$ & 7.5 &  20.3 \\
+ DAS - Ball-Query $\ell_0$ & 7.9 & 19.3\\
\noalign{\global\arrayrulewidth1pt}\hline\noalign{\global\arrayrulewidth0.4pt}
\end{tabular}
\end{minipage}
\hfill
\begin{minipage}{0.45\linewidth}
\centering
\caption{ Ablation Study on DAS for Different Models.}
\label{tb:DWS different architectures}
\begin{tabular}{lcc}
\noalign{\global\arrayrulewidth1pt}\hline\noalign{\global\arrayrulewidth0.4pt}
Model (\%) $\downarrow$ & $\text{ER}_\text{clean}$ & $\text{ER}_\text{cor}$ \\
\hline
CurveNet & 6.9  & 24.3 \\
\rowcolor{gray!20}+ DAS & \textbf{6.6} & \textbf{23.1} \\
PointMLP & 7.1 & 32.2 \\
\rowcolor{gray!20}+ DAS & \textbf{7.1} & \textbf{28.0} \\
\noalign{\global\arrayrulewidth1pt}\hline\noalign{\global\arrayrulewidth0.4pt}
\end{tabular}
\end{minipage}

\end{table}

\vspace{-0.2cm}
\section{Discussion and Conclusion}
\vspace{-0.2cm}

Robustness to data corruption is a crucial aspect in the field of 3D point cloud recognition. In real-world scenarios, the data acquired through sensors and other means often contain noise, occlusions, or other forms of corruption which can severely impede the performance of recognition systems. Unlike traditional 2D image data, 3D point clouds encapsulate geometric and spatial information which, when corrupted, can lead to a significant loss of critical information required for accurate recognition. The integrity of this data is fundamental, especially in critical applications like autonomous navigation, robotics, and real-time monitoring systems where a slight misinterpretation of the data could lead to catastrophic results. Furthermore, robustness to data corruption ensures the stability and reliability of 3D recognition systems across varying conditions and environments, making them more versatile and practical for real-world applications. Additionally, by ensuring a system’s robustness to data corruption, we pave the way for more accurate and reliable models that can better generalize across different scenarios. In essence, addressing the challenges posed by data corruption in 3D point cloud recognition not only significantly enhances the performance of recognition systems but also broadens the scope and applicability of 3D recognition technologies in real-life scenarios.

In conclusion, our research leverage the unique set property of point cloud data, culminating in the development of a pioneering critical subset identification (CSI) method designed to enhance recognition robustness against data corruption. Central to our CSI approach are two key innovations: density-aware sampling (DAS) and self-entropy minimization (SEM). DAS optimizes robust anchor point sampling by considering local density, while SEM, applied during the training phase, emphasizes the most crucial point-to-point attention. Our empirical assessments underscore the efficacy of CSI, with error rates reduced to 18.4\% on ModelNet40-C and 16.3\% on PointCloud-C. These results represent substantial gains, outpacing SOTA methods by margins of 5.2\% and 4.2\% on the respective benchmarks.

\textbf{Limitations and Future Work}.  Our proposed CSI demonstrates a notable improvement over standard training techniques. However, when combined with data augmentation, the margin of improvement is relatively modest. This can be attributed to the inherent complexity of data augmentation methods in 3D point clouds, which often entail the blending of point clouds from different classes. This amalgamation creates a level of ambiguity that makes the identification of a critical subset in the augmented point cloud challenging. The efficacy of DAS is primarily observed in its resilience to noise corruption. While our method offers an empirical solution to enhancing corruption robustness, a certified solution remains unexplored, presenting a potentially promising avenue for future research.

\clearpage
\bibliography{iclr2021_conference}

\begin{thebibliography}{47}
\providecommand{\natexlab}[1]{#1}
\providecommand{\url}[1]{\texttt{#1}}
\expandafter\ifx\csname urlstyle\endcsname\relax
  \providecommand{\doi}[1]{doi: #1}\else
  \providecommand{\doi}{doi: \begingroup \urlstyle{rm}\Url}\fi

\bibitem[Burks et~al.()Burks, Shannon, and Weaver]{shannon2001mathematical}
Arthur~W. Burks, Claude~E. Shannon, and Warren Weaver.
\newblock The mathematical theory of communication.
\newblock \emph{The Philosophical Review}, pp.\  398.
\newblock \doi{10.2307/2181879}.

\bibitem[Charles et~al.(2017)Charles, Su, Kaichun, and Guibas]{Charles_Su_Kaichun_Guibas_2017}
R.~Qi Charles, Hao Su, Mo~Kaichun, and Leonidas~J. Guibas.
\newblock Pointnet: Deep learning on point sets for 3d classification and segmentation.
\newblock In \emph{2017 Proceedings of the IEEE Conference on Computer Vision and Pattern Recognition (CVPR)}, Jul 2017.
\newblock \doi{10.1109/cvpr.2017.16}.

\bibitem[Chen et~al.(2017)Chen, Ma, Wan, Li, and Xia]{DBLP:conf/cvpr/ChenMWLX17}
Xiaozhi Chen, Huimin Ma, Ji~Wan, Bo~Li, and Tian Xia.
\newblock Multi-view 3d object detection network for autonomous driving.
\newblock In \emph{2017 Proceedings of the {IEEE} Conference on Computer Vision and Pattern Recognition (CVPR)}, pp.\  6526--6534. {IEEE} Computer Society, 2017.
\newblock \doi{10.1109/CVPR.2017.691}.

\bibitem[Chen et~al.(2020)Chen, Hu, Gavves, Mensink, Mettes, Yang, and Snoek]{DBLP:conf/eccv/ChenHGMMYS20}
Yunlu Chen, Vincent~Tao Hu, Efstratios Gavves, Thomas Mensink, Pascal Mettes, Pengwan Yang, and Cees G.~M. Snoek.
\newblock Pointmixup: Augmentation for point clouds.
\newblock In Andrea Vedaldi, Horst Bischof, Thomas Brox, and Jan{-}Michael Frahm (eds.), \emph{Computer Vision - {ECCV} 2020 - 16th European Conference, Glasgow, UK, August 23-28, 2020, Proceedings, Part {III}}, volume 12348 of \emph{Lecture Notes in Computer Science}, pp.\  330--345. Springer, 2020.
\newblock \doi{10.1007/978-3-030-58580-8\_20}.

\bibitem[Dong et~al.(2020)Dong, Chen, Zhou, Hua, Zhang, and Yu]{Dong_Chen_Zhou_Hua_Zhang_Yu_2020}
Xiaoyi Dong, Dongdong Chen, Hang Zhou, Gang Hua, Weiming Zhang, and Nenghai Yu.
\newblock Self-robust 3d point recognition via gather-vector guidance.
\newblock In \emph{2020 Proceedings of the IEEE/CVF Conference on Computer Vision and Pattern Recognition (CVPR)}, Jun 2020.
\newblock \doi{10.1109/cvpr42600.2020.01153}.

\bibitem[Garrido et~al.(2021)Garrido, Rodrigues, Augusto~Sousa, Jacob, and Castro~Silva]{Garrido_Rodrigues_Augusto_Sousa_Jacob_Castro_Silva_2021}
Daniel Garrido, Rui Rodrigues, A.~Augusto~Sousa, Joao Jacob, and Daniel Castro~Silva.
\newblock Point cloud interaction and manipulation in virtual reality.
\newblock In \emph{2021 Proceedings of the International Conference on Artificial Intelligence and Virtual Reality (AIVR)}, Jul 2021.
\newblock \doi{10.1145/3480433.3480437}.

\bibitem[Goyal et~al.(2021)Goyal, Law, Liu, Newell, and Deng]{goyal2021revisiting}
Ankit Goyal, Hei Law, Bowei Liu, Alejandro Newell, and Jia Deng.
\newblock Revisiting point cloud shape classification with a simple and effective baseline.
\newblock In \emph{2021 Proceedings of the 38th International Conference on Machine Learning (ICML)}, pp.\  3809--3820. PMLR, 2021.

\bibitem[Goyal et~al.(2022)Goyal, Sun, Raghunathan, and Kolter]{goyal2022test}
Sachin Goyal, Mingjie Sun, Aditi Raghunathan, and J~Zico Kolter.
\newblock Test time adaptation via conjugate pseudo-labels.
\newblock \emph{2022 Advances in Neural Information Processing Systems: Annual Conference on Neural Information Processing Systems (NeurIPS)}, 35:\penalty0 6204--6218, 2022.

\bibitem[Guo et~al.(2021)Guo, Cai, Liu, Mu, Martin, and Hu]{guo2021pct}
Meng-Hao Guo, Jun-Xiong Cai, Zheng-Ning Liu, Tai-Jiang Mu, Ralph~R. Martin, and Shi-Min Hu.
\newblock Pct: Point cloud transformer.
\newblock \emph{Computational Visual Media}, pp.\  187–199, Jun 2021.
\newblock \doi{10.1007/s41095-021-0229-5}.

\bibitem[Kim et~al.(2021)Kim, Lee, Hwang, Lee, Hwang, and Kim]{DBLP:conf/iccv/KimLHLHK21}
Sihyeon Kim, Sanghyeok Lee, Dasol Hwang, Jaewon Lee, Seong~Jae Hwang, and Hyunwoo~J. Kim.
\newblock Point cloud augmentation with weighted local transformations.
\newblock In \emph{2021 Proceedings of the {IEEE} International Conference on Computer Vision (ICCV)}, pp.\  528--537. {IEEE}, 2021.
\newblock \doi{10.1109/ICCV48922.2021.00059}.

\bibitem[Kingma \& Ba(2014)Kingma and Ba]{Kingma_Ba_2014}
DiederikP. Kingma and Jimmy Ba.
\newblock Adam: A method for stochastic optimization.
\newblock \emph{arXiv: Learning}, Dec 2014.

\bibitem[Landrieu \& Simonovsky(2018)Landrieu and Simonovsky]{Landrieu_Simonovsky_2018}
Loic Landrieu and Martin Simonovsky.
\newblock Large-scale point cloud semantic segmentation with superpoint graphs.
\newblock In \emph{2018 Proceedings of the IEEE/CVF Conference on Computer Vision and Pattern Recognition (CVPR)}, Jun 2018.
\newblock \doi{10.1109/cvpr.2018.00479}.

\bibitem[Lee et~al.(2021{\natexlab{a}})Lee, Lee, Lee, Lee, Lee, Woo, and Lee]{DBLP:conf/cvpr/LeeLLLLWL21}
Dogyoon Lee, Jaeha Lee, Junhyeop Lee, Hyeongmin Lee, Minhyeok Lee, Sungmin Woo, and Sangyoun Lee.
\newblock Regularization strategy for point cloud via rigidly mixed sample.
\newblock In \emph{2021 Proceedings of the {IEEE} Conference on Computer Vision and Pattern Recognition (2021)}, pp.\  15900--15909. Computer Vision Foundation / {IEEE}, 2021{\natexlab{a}}.
\newblock \doi{10.1109/CVPR46437.2021.01564}.

\bibitem[Lee et~al.(2021{\natexlab{b}})Lee, Lee, Lee, Lee, Lee, Woo, and Lee]{Lee_Lee_Lee_Lee_Lee_Woo_Lee_2021}
Dogyoon Lee, Jaeha Lee, Junhyeop Lee, Hyeongmin Lee, Minhyeok Lee, Sungmin Woo, and Sangyoun Lee.
\newblock Regularization strategy for point cloud via rigidly mixed sample.
\newblock In \emph{2021 Proceedings of the IEEE/CVF Conference on Computer Vision and Pattern Recognition (CVPR)}, Jun 2021{\natexlab{b}}.
\newblock \doi{10.1109/cvpr46437.2021.01564}.

\bibitem[Liu et~al.(2021)Liu, Jia, and Gong]{DBLP:journals/corr/abs-2103-03046}
Hongbin Liu, Jinyuan Jia, and Neil~Zhenqiang Gong.
\newblock Pointguard: Provably robust 3d point cloud classification.
\newblock In \emph{2021 Proceedings of the {IEEE} Conference on Computer Vision and Pattern Recognition (CVPR)}, pp.\  6186--6195. Computer Vision Foundation / {IEEE}, 2021.
\newblock \doi{10.1109/CVPR46437.2021.00612}.

\bibitem[Liu et~al.(2019)Liu, Fan, Xiang, and Pan]{Liu_Fan_Xiang_Pan_2019}
Yongcheng Liu, Bin Fan, Shiming Xiang, and Chunhong Pan.
\newblock Relation-shape convolutional neural network for point cloud analysis.
\newblock In \emph{2019 Proceedings of the IEEE/CVF Conference on Computer Vision and Pattern Recognition (CVPR)}, Jun 2019.
\newblock \doi{10.1109/cvpr.2019.00910}.

\bibitem[Ma et~al.(2022)Ma, Qin, You, Ran, and Fu]{DBLP:conf/iclr/MaQYR022}
Xu~Ma, Can Qin, Haoxuan You, Haoxi Ran, and Yun Fu.
\newblock Rethinking network design and local geometry in point cloud: {A} simple residual {MLP} framework.
\newblock In \emph{2022 The Tenth International Conference on Learning Representations (ICLR)}. OpenReview.net, 2022.

\bibitem[Muzahid et~al.(2021)Muzahid, Wan, Sohel, Wu, and Hou]{Muzahid_Wan_Sohel_Wu_Hou_2021}
A.~A.~M. Muzahid, Wanggen Wan, Ferdous Sohel, Lianyao Wu, and Li~Hou.
\newblock Curvenet: Curvature-based multitask learning deep networks for 3d object recognition.
\newblock \emph{IEEE/CAA Journal of Automatica Sinica}, pp.\  1177–1187, Jun 2021.
\newblock \doi{10.1109/jas.2020.1003324}.

\bibitem[Paszke et~al.(2019)Paszke, Gross, Massa, Lerer, Bradbury, Chanan, Killeen, Lin, Gimelshein, Antiga, Desmaison, K{\"{o}}pf, Yang, DeVito, Raison, Tejani, Chilamkurthy, Steiner, Fang, Bai, and Chintala]{DBLP:conf/nips/PaszkeGMLBCKLGA19}
Adam Paszke, Sam Gross, Francisco Massa, Adam Lerer, James Bradbury, Gregory Chanan, Trevor Killeen, Zeming Lin, Natalia Gimelshein, Luca Antiga, Alban Desmaison, Andreas K{\"{o}}pf, Edward~Z. Yang, Zachary DeVito, Martin Raison, Alykhan Tejani, Sasank Chilamkurthy, Benoit Steiner, Lu~Fang, Junjie Bai, and Soumith Chintala.
\newblock Pytorch: An imperative style, high-performance deep learning library.
\newblock In Hanna~M. Wallach, Hugo Larochelle, Alina Beygelzimer, Florence d'Alch{\'{e}}{-}Buc, Emily~B. Fox, and Roman Garnett (eds.), \emph{2019 Advances in Neural Information Processing Systems: Annual Conference on Neural Information Processing Systems (NeurIPS)}, pp.\  8024--8035, 2019.

\bibitem[Qi et~al.(2017)Qi, Yi, Su, and Guibas]{qi2017pointnet++}
Charles~Ruizhongtai Qi, Li~Yi, Hao Su, and Leonidas~J Guibas.
\newblock Pointnet++: Deep hierarchical feature learning on point sets in a metric space.
\newblock \emph{Advances in neural information processing systems}, 30, 2017.

\bibitem[Ren et~al.(2022)Ren, Pan, and Liu]{ren2022benchmarking}
Jiawei Ren, Liang Pan, and Ziwei Liu.
\newblock Benchmarking and analyzing point cloud classification under corruptions.
\newblock In \emph{Proceedings of the 39th International Conference on Machine Learning, {ICML} 2022}, pp.\  18559--18575. PMLR, 2022.

\bibitem[Shen et~al.(2018)Shen, Feng, Yang, and Tian]{DBLP:conf/cvpr/ShenFYT18}
Yiru Shen, Chen Feng, Yaoqing Yang, and Dong Tian.
\newblock Mining point cloud local structures by kernel correlation and graph pooling.
\newblock In \emph{2018 Proceedings of the {IEEE} Conference on Computer Vision and Pattern Recognition (CVPR)}, pp.\  4548--4557. Computer Vision Foundation / {IEEE} Computer Society, 2018.
\newblock \doi{10.1109/CVPR.2018.00478}.

\bibitem[Simonovsky \& Komodakis(2017)Simonovsky and Komodakis]{DBLP:conf/cvpr/SimonovskyK17}
Martin Simonovsky and Nikos Komodakis.
\newblock Dynamic edge-conditioned filters in convolutional neural networks on graphs.
\newblock In \emph{2017 Proceedings of the {IEEE} Conference on Computer Vision and Pattern Recognition (CVPR)}, pp.\  29--38. {IEEE} Computer Society, 2017.
\newblock \doi{10.1109/CVPR.2017.11}.

\bibitem[Sun et~al.(2020)Sun, Koenig, Cao, Chen, and Mao]{sun2020adversarial}
Jiachen Sun, Karl Koenig, Yulong Cao, Qi~Alfred Chen, and Z~Morley Mao.
\newblock On adversarial robustness of 3d point cloud classification under adaptive attacks.
\newblock \emph{arXiv preprint arXiv:2011.11922}, 2020.

\bibitem[Sun et~al.(2021)Sun, Cao, Choy, Yu, Anandkumar, Mao, and Xiao]{Sun_Cao_Choy_Yu_Anandkumar_Mao_Xiao_2021}
Jiachen Sun, Yulong Cao, Christopher Choy, Zhiding Yu, Animashree Anandkumar, ZhuoqingMorley Mao, and Chaowei Xiao.
\newblock Adversarially robust 3d point cloud recognition using self-supervisions.
\newblock \emph{2021 Advances in Neural Information Processing Systems: Annual Conference on Neural Information Processing Systems (NeurIPS)}, Dec 2021.

\bibitem[Sun et~al.(2022)Sun, Zhang, Kailkhura, Yu, Xiao, and Mao]{sun2022benchmarking}
Jiachen Sun, Qingzhao Zhang, Bhavya Kailkhura, Zhiding Yu, Chaowei Xiao, and Z~Morley Mao.
\newblock Benchmarking robustness of 3d point cloud recognition against common corruptions.
\newblock \emph{arXiv preprint arXiv:2201.12296}, 2022.

\bibitem[Sun et~al.(2023)Sun, Wang, Nie, Yu, Mao, and Xiao]{sun2023critical}
Jiachen Sun, Jiongxiao Wang, Weili Nie, Zhiding Yu, Zhuoqing Mao, and Chaowei Xiao.
\newblock A critical revisit of adversarial robustness in 3d point cloud recognition with diffusion-driven purification.
\newblock In Andreas Krause, Emma Brunskill, Kyunghyun Cho, Barbara Engelhardt, Sivan Sabato, and Jonathan Scarlett (eds.), \emph{2023 Proceedings of the 39th International Conference on Machine Learning (ICML)}, volume 202 of \emph{Proceedings of Machine Learning Research}, pp.\  33100--33114. {PMLR}, 2023.

\bibitem[Taghanaki et~al.(2020)Taghanaki, Luo, Zhang, Wang, Jayaraman, and Jatavallabhula]{DBLP:journals/corr/abs-2011-11572}
Saeid~Asgari Taghanaki, Jieliang Luo, Ran Zhang, Ye~Wang, Pradeep~Kumar Jayaraman, and Krishna~Murthy Jatavallabhula.
\newblock Robustpointset: {A} dataset for benchmarking robustness of point cloud classifiers.
\newblock \emph{CoRR}, abs/2011.11572, 2020.

\bibitem[Tashi et~al.(2019)Tashi, Ullah, and Kubo]{Tashi_Ullah_Kubo_2019}
Tashi, AMM~Sharif Ullah, and Akihiko Kubo.
\newblock Geometric modeling and 3d printing using recursively generated point cloud.
\newblock \emph{Mathematical and Computational Applications}, 24\penalty0 (3):\penalty0 83, Sep 2019.
\newblock \doi{10.3390/mca24030083}.

\bibitem[Vaswani et~al.(2017)Vaswani, Shazeer, Parmar, Uszkoreit, Jones, Gomez, Kaiser, and Polosukhin]{vaswani2017attention}
Ashish Vaswani, Noam Shazeer, Niki Parmar, Jakob Uszkoreit, Llion Jones, AidanN. Gomez, Lukasz Kaiser, and Illia Polosukhin.
\newblock Attention is all you need.
\newblock \emph{Neural Information Processing Systems}, Jun 2017.

\bibitem[Wang et~al.(2021{\natexlab{a}})Wang, Shelhamer, Liu, Olshausen, and Darrell]{DBLP:conf/iclr/WangSLOD21}
Dequan Wang, Evan Shelhamer, Shaoteng Liu, Bruno~A. Olshausen, and Trevor Darrell.
\newblock Tent: Fully test-time adaptation by entropy minimization.
\newblock In \emph{2021 the Ninth International Conference on Learning Representations (ICLR)}. OpenReview.net, 2021{\natexlab{a}}.

\bibitem[Wang et~al.(2021{\natexlab{b}})Wang, Liu, Yue, Lasenby, and Kusner]{Wang_Liu_Yue_Lasenby_Kusner_2021}
Hanchen Wang, Qi~Liu, Xiangyu Yue, Joan Lasenby, and Matt~J. Kusner.
\newblock Unsupervised point cloud pre-training via occlusion completion.
\newblock In \emph{2021 Proceedings of the IEEE/CVF International Conference on Computer Vision (ICCV)}, Oct 2021{\natexlab{b}}.
\newblock \doi{10.1109/iccv48922.2021.00964}.

\bibitem[Wang et~al.(2019)Wang, Sun, Liu, Sarma, Bronstein, and Solomon]{wang2019dynamic}
Yue Wang, Yongbin Sun, Ziwei Liu, Sanjay~E. Sarma, Michael~M. Bronstein, and Justin~M. Solomon.
\newblock Dynamic graph cnn for learning on point clouds.
\newblock \emph{ACM Transactions on Graphics}, pp.\  1–12, Oct 2019.
\newblock \doi{10.1145/3326362}.

\bibitem[Wu et~al.(2019)Wu, Zhou, Zhao, Yue, and Keutzer]{Wu_Zhou_Zhao_Yue_Keutzer_2019}
Bichen Wu, Xuanyu Zhou, Sicheng Zhao, Xiangyu Yue, and Kurt Keutzer.
\newblock Squeezesegv2: Improved model structure and unsupervised domain adaptation for road-object segmentation from a lidar point cloud.
\newblock In \emph{2019 Proceedings of the International Conference on Robotics and Automation (ICRA)}, May 2019.
\newblock \doi{10.1109/icra.2019.8793495}.

\bibitem[Wu et~al.(2022)Wu, Lao, Jiang, Liu, and Zhao]{DBLP:conf/nips/0002LJLZ22}
Xiaoyang Wu, Yixing Lao, Li~Jiang, Xihui Liu, and Hengshuang Zhao.
\newblock Point transformer {V2:} grouped vector attention and partition-based pooling.
\newblock In \emph{2022 Advances in Neural Information Processing Systems: Annual Conference on Neural Information Processing Systems (NeurIPS)}, 2022.

\bibitem[Wu et~al.(2015)Wu, Song, Khosla, Yu, Zhang, Tang, and Xiao]{wu20153d}
Zhirong Wu, Shuran Song, Aditya Khosla, Fisher Yu, Linguang Zhang, Xiaoou Tang, and Jianxiong Xiao.
\newblock 3d shapenets: A deep representation for volumetric shapes.
\newblock In \emph{2015 Proceedings of the IEEE Conference on Computer Vision and Pattern Recognition (CVPR)}, Jun 2015.
\newblock \doi{10.1109/cvpr.2015.7298801}.

\bibitem[Xiao \& Wachs(2021)Xiao and Wachs]{Xiao_Wachs_2021}
Chenxi Xiao and Juan Wachs.
\newblock Triangle-net: Towards robustness in point cloud learning.
\newblock In \emph{2021 Proceedings of the IEEE Winter Conference on Applications of Computer Vision (WACV)}, Jan 2021.
\newblock \doi{10.1109/wacv48630.2021.00087}.

\bibitem[Xu et~al.(2021)Xu, Ding, Zhao, and Qi]{Xu_Ding_Zhao_Qi_2021}
Mutian Xu, Runyu Ding, Hengshuang Zhao, and Xiaojuan Qi.
\newblock Paconv: Position adaptive convolution with dynamic kernel assembling on point clouds.
\newblock In \emph{2021 Proceedings of the IEEE/CVF Conference on Computer Vision and Pattern Recognition (CVPR)}, Jun 2021.
\newblock \doi{10.1109/cvpr46437.2021.00319}.

\bibitem[Xu et~al.(2022)Xu, Zhang, Zhou, Xu, Qi, and Qiao]{Xu_Zhang_Zhou_Xu_Qi_Qiao_2022}
Mutian Xu, Junhao Zhang, Zhipeng Zhou, Mingye Xu, Xiaojuan Qi, and Yu~Qiao.
\newblock Learning geometry-disentangled representation for complementary understanding of 3d object point cloud.
\newblock \emph{2022 Proceedings of the AAAI Conference on Artificial Intelligence (AAAI)}, pp.\  3056–3064, Sep 2022.
\newblock \doi{10.1609/aaai.v35i4.16414}.

\bibitem[Yan et~al.(2020)Yan, Zheng, Li, Wang, and Cui]{Yan_Zheng_Li_Wang_Cui_2020}
Xu~Yan, Chaoda Zheng, Zhen Li, Sheng Wang, and Shuguang Cui.
\newblock Pointasnl: Robust point clouds processing using nonlocal neural networks with adaptive sampling.
\newblock In \emph{2020 Proceedings of the IEEE/CVF Conference on Computer Vision and Pattern Recognition (CVPR)}, Jun 2020.
\newblock \doi{10.1109/cvpr42600.2020.00563}.

\bibitem[Yu et~al.(2022)Yu, Tang, Rao, Huang, Zhou, and Lu]{yu2021pointbert}
Xumin Yu, Lulu Tang, Yongming Rao, Tiejun Huang, Jie Zhou, and Jiwen Lu.
\newblock Point-bert: Pre-training 3d point cloud transformers with masked point modeling.
\newblock In \emph{2022 Proceedings of the IEEE Conference on Computer Vision and Pattern Recognition (CVPR)}, 2022.

\bibitem[Yue et~al.(2018)Yue, Wu, Seshia, Keutzer, and Sangiovanni{-}Vincentelli]{DBLP:conf/mir/YueWSKS18}
Xiangyu Yue, Bichen Wu, Sanjit~A. Seshia, Kurt Keutzer, and Alberto~L. Sangiovanni{-}Vincentelli.
\newblock A lidar point cloud generator: from a virtual world to autonomous driving.
\newblock In Kiyoharu Aizawa, Michael~S. Lew, and Shin'ichi Satoh (eds.), \emph{2018 Proceedings of the 2018 {ACM} on International Conference on Multimedia Retrieval (ICMR)}, pp.\  458--464. {ACM}, 2018.
\newblock \doi{10.1145/3206025.3206080}.

\bibitem[Yun et~al.(2019)Yun, Han, Chun, Oh, Yoo, and Choe]{Yun_Han_Chun_Oh_Yoo_Choe_2019}
Sangdoo Yun, Dongyoon Han, Sanghyuk Chun, Seong~Joon Oh, Youngjoon Yoo, and Junsuk Choe.
\newblock Cutmix: Regularization strategy to train strong classifiers with localizable features.
\newblock In \emph{2019 Proceedings of the {IEEE} International Conference on Computer Vision (ICCV)}, Oct 2019.
\newblock \doi{10.1109/iccv.2019.00612}.

\bibitem[Zhang et~al.(2018)Zhang, Ciss{\'{e}}, Dauphin, and Lopez{-}Paz]{DBLP:conf/iclr/ZhangCDL18}
Hongyi Zhang, Moustapha Ciss{\'{e}}, Yann~N. Dauphin, and David Lopez{-}Paz.
\newblock mixup: Beyond empirical risk minimization.
\newblock In \emph{2018 the Sith International Conference on Learning Representations (ICLR)}. OpenReview.net, 2018.

\bibitem[Zhang et~al.(2022{\natexlab{a}})Zhang, Chen, Ouyang, Liu, Zhu, Chen, Meng, and Wu]{Zhang_Chen_Ouyang_Liu_Zhu_Chen_Meng_Wu_2022}
Jinlai Zhang, Lyujie Chen, Bo~Ouyang, Binbin Liu, Jihong Zhu, Yujin Chen, Yanmei Meng, and Danfeng Wu.
\newblock Pointcutmix: Regularization strategy for point cloud classification.
\newblock \emph{Neurocomputing}, pp.\  58–67, Sep 2022{\natexlab{a}}.
\newblock \doi{10.1016/j.neucom.2022.07.049}.

\bibitem[Zhang et~al.(2022{\natexlab{b}})Zhang, Levine, and Finn]{DBLP:conf/nips/ZhangLF22}
Marvin Zhang, Sergey Levine, and Chelsea Finn.
\newblock {MEMO:} test time robustness via adaptation and augmentation.
\newblock In \emph{2022 Advances in Neural Information Processing Systems 32: Annual Conference on Neural Information Processing Systems (NeurIPS)}, 2022{\natexlab{b}}.

\bibitem[Zhao et~al.(2021)Zhao, Jiang, Jia, Torr, and Koltun]{Zhao_Jiang_Jia_Torr_Koltun_2021}
Hengshuang Zhao, Li~Jiang, Jiaya Jia, Philip Torr, and Vladlen Koltun.
\newblock Point transformer.
\newblock In \emph{2021 IEEE/CVF International Conference on Computer Vision (ICCV)}, Oct 2021.
\newblock \doi{10.1109/iccv48922.2021.01595}.

\end{thebibliography}
\bibliographystyle{iclr2021_conference}

\appendix
\setcounter{table}{0}
\setcounter{figure}{0}
\newpage
\section{Evaluation Details}
We report the detailed evaluation results in Tables~\ref{modelnet40-c} and~\ref{tab:PointCloud-C}.
\label{ap:eval}
\begin{table}[h!]
\centering
\caption{ Error Rates of Different Model Architectures on ModelNet40-C and ModelNet40 with Standard Training.}
\label{modelnet40-c}
\resizebox{\linewidth}{!}{
\begin{tabular}{lccccccc}
\noalign{\global\arrayrulewidth1pt}\hline\noalign{\global\arrayrulewidth0.4pt}

Type (\%) $\downarrow$ & PointNet & PointNet++ & DGCNN & RSCNN & SimpleView & PCT & PCT+CSI \\
\hline

Occlusion & 52.3 & 54.7 & 59.2 & 51.8 & 55.5 & 56.6 & 56.2 \\
LiDAR & 54.9 & 66.5 & 81.0 & 68.4 & 82.2 & 76.7 & 60.2 \\
Density Inc. & 10.5 & 16.0 & 14.1 & 16.8 & 13.7 & 11.8 & 12.4 \\
Density Dec. & 11.6 & 10.0 & 17.3 & 13.2 & 17.2 & 14.3 & 12.7 \\
Cutout & 12.0 & 10.7 & 15.4 & 13.8 & 20.1 & 14.5 & 12.0 \\
Uniform & 12.4 & 20.4 & 14.6 & 24.6 & 14.5 & 12.1 & 10.8 \\
Gaussian & 14.4 & 16.4 & 16.6 & 18.3 & 14.2 & 13.9 & 10.7 \\
Impulse & 29.1 & 35.1 & 24.9 & 46.2 & 24.6 & 39.1 & 12.3 \\
Upsampling & 14.0 & 17.2 & 19.1 & 18.3 & 17.7 & 57.9 & 11.6 \\
Background & 93.6 & 18.6 & 19.1 & 29.2 & 46.8 & 18.1 & 11.2 \\
Rotation & 36.8 & 27.6 & 12.1 & 17.0 & 30.7 & 11.5 & 17.4 \\

Shear & 25.4 & 13.4 & 13.1 & 18.1 & 18.5 & 12.4 & 12.0 \\

FFD & 21.3 & 15.2 & 14.5 & 19.2 & 17.0 & 13.0 & 12.1 \\

RBF & 18.6 & 16.4 & 14.0 & 18.6 & 17.9 & 12.6 & 12.4 \\

Inv. RBF & 17.8 & 15.4  & 14.5 & 18.6 & 17.2 & 12.6 & 11.7
\\
\hline
$\text{ER}_\text{cor}$ & 28.3 & 23.6 & 25.9 & 26.2 & 27.2 & 25.5 & 18.4 \\
$\text{ER}_\text{clean}$ & 9.3 & 7.0 & 7.4 & 7.7 & 6.1 & 7.1 & 7.3 \\
\noalign{\global\arrayrulewidth1pt}\hline\noalign{\global\arrayrulewidth0.4pt}
\end{tabular}
}
\end{table}

\begin{table*}[!]
\centering
\caption{Full Results of the Overall Accuracy (OA). mOA: average OA over all corruptions.}
\label{tab:PointCloud-C}
\resizebox{\linewidth}{!}{

\begin{tabular}{lccccccccc}
\noalign{\global\arrayrulewidth1pt}\hline\noalign{\global\arrayrulewidth0.4pt}
{Model (\%) $\uparrow$} &                                       Clean  &                                  mOA &                                       Scale &                                      Jitter &                                      Drop-G &                                      Drop-L &                                       Add-G &                                    Add-L &                                      Rotate \\
\midrule
DGCNN                           &               0.926 &            0.764 &               0.906 &               0.684 &               0.752 &               0.793 &               0.705 &            0.725 &               0.785 \\
PointNet                       &               0.907 &            0.658 &               0.881 &               0.797 &               0.876 &               0.778 &               0.121 &            0.562 &               0.591 \\
PointNet++                     &               0.930 &            0.751 &               0.918 &               0.628 &               0.841 &               0.627 &               0.819 &            0.727 &               0.698 \\
RSCNN                           &               0.923 &            0.739 &               0.899 &               0.630 &               0.800 &               0.686 &               0.790 &            0.683 &               0.682 \\
SimpleView                      &     {0.939} &            0.757 &               0.918 &               0.774 &               0.692 &               0.719 &               0.710 &            0.768 &               0.717 \\
GDANet                          &               0.934 &            0.789 &     {0.922} &               0.735 &               0.803 &               0.815 &               0.743 &            0.715 &               0.789 \\
CurveNet                        &  {0.938} &            0.779 &               0.918 &               0.771 &               0.824 &               0.788 &               0.603 &            0.725 &               0.826 \\
PAConv                          &               0.936 &            0.730 &               0.915 &               0.537 &               0.752 &               0.792 &               0.680 &            0.643 &               0.792 \\
RPC                       &               0.930 &            0.795 &  {0.921} &               0.718 &               0.878 &               0.835 &               0.726 &            0.722 &               0.768 \\
PCT                             &               0.930 &            0.781 &               0.918 &               0.725 &               0.869 &               0.793 &               0.770 &            0.619 &               0.776 \\

PCT+CSI & 0.927 & 0.837 & 0.895 & 0.848 & 0.861 & 0.816 & 0.905 & 0.747 & 0.787 \\
\midrule 
DGCNN+OcCo                      &               0.922 &            0.766 &               0.849 &               0.794 &               0.776 &               0.785 &               0.574 &            0.767 &               0.820 \\
Point-BERT                      &               0.922 &            0.693 &               0.912 &               0.602 &               0.829 &               0.762 &               0.430 &            0.604 &               0.715 \\
\midrule PN2+PointMixUp                     &               0.915 &            0.785 &               0.843 &               0.775 &               0.801 &               0.625 &               0.865 &            0.831 &               0.757 \\
DGCNN+PW                 &               0.926 &            0.809 &               0.913 &               0.727 &               0.755 &               0.819 &               0.762 &            0.790 &               0.897 \\
DGCNN+RSMix                     &               0.930 &            0.839 &               0.876 &               0.724 &               0.838 &               0.878 &     {0.917} &            0.827 &               0.813 \\
DGCNN+WOLFMix                   &               0.932 &  {0.871} &               0.907 &               0.774 &               0.827 &               0.881 &  {0.916} &  {0.886} &  {0.903} \\
\midrule
PointNet+WOLFMix                &               0.884 &            0.743 &               0.801 &  {0.850} &               0.857 &               0.776 &               0.343 &            0.807 &               0.768 \\
PCT+WOLFMix                     &               0.934 &            0.873 &               0.906 &               0.730 &  {0.906} &               0.898 &               0.912 &            0.861 &  {0.895} \\
GDANet+WOLFMix                     &               0.934 &  \textrm{0.871} &               0.915 &               0.721 &               0.868 &     {0.886} &               0.910 &  {0.886} &     {0.912} \\
RPC+WOLFMix               &               0.933 &            0.865 &               0.905 &               0.694 &               0.895 &               0.894 &               0.902 &            0.868 &               0.897 \\
PCT+CSI+PointCutMix-R & 0.928 & 0.871 & 0.887 & 0.885 & 0.887 & 0.841 & 0.918 & 0.878 & 0.837 \\
\noalign{\global\arrayrulewidth1pt}\hline\noalign{\global\arrayrulewidth0.4pt}
\end{tabular}
}
\end{table*}


\end{document}